%% file: main.tex
\documentclass[letterpaper, 10 pt, conference]{ieeeconf}  

\IEEEoverridecommandlockouts                              

\usepackage{graphics} 
\usepackage{epsfig} 
\usepackage{mathptmx} 
\usepackage{times} 
\usepackage{pifont}
\usepackage{amsmath} 
\usepackage{amssymb}  
\usepackage{siunitx}
\usepackage{academicons}
\usepackage{fontawesome5}
\usepackage{marvosym}
\usepackage{xspace}
\usepackage{graphicx}
\usepackage{amsmath}
\usepackage{amssymb}
\usepackage{booktabs}
\usepackage{makecell}
\usepackage{multirow}
\usepackage{cellspace, tabularx}
\usepackage{colortbl}
\usepackage{bigstrut}
\usepackage{bbm}
\usepackage{ifsym}
\usepackage{csquotes}
\usepackage{algorithm}
\usepackage{algpseudocode}
\usepackage{hyperref}
\hypersetup{
  colorlinks = true, 
  urlcolor = blue, 
  linkcolor = blue, 
  citecolor = blue, 
  breaklinks = False,
}

\algrenewcommand\algorithmicrequire{\textbf{Input:}}
\algrenewcommand\algorithmicensure{\textbf{Output:}}
\usepackage{textcomp, gensymb}
\usepackage[english]{babel} 
\usepackage{blindtext}
\DeclareUnicodeCharacter{0301}{'}

\usepackage[backend=biber,
            hyperref=true,
            url=false,
            isbn=false,
            doi=false,
            backref=false,
            style=ieee,
            citestyle=numeric-comp,
            sorting=nyt,
            block=none]{biblatex}
\usepackage[font=footnotesize]{caption}

\usepackage[dvipsnames]{xcolor}
\usepackage{multicol}

\title{\LARGE \bf
SPOT: SE(3) Pose Trajectory Diffusion for Object-Centric Manipulation
}

\author{
    Cheng-Chun Hsu\textsuperscript{1,2,*}, Bowen Wen\textsuperscript{1,\Letter}, Jie Xu$^{1}$, Yashraj Narang$^{1}$, Xiaolong Wang$^{1,3}$,
    \\
    Yuke Zhu$^{1,2}$, Joydeep Biswas$^{1,2}$, Stan Birchfield$^{1}$
    \thanks{
        $^{1}$ NVIDIA.
        $^{2}$ University of Texas at Austin.
        $^{3}$ University of California San Diego. $*$Work done during internship at NVIDIA. \Letter Corresponding author.}
}

\renewcommand{\bibfont}{\small}
\newenvironment{myitem}{\begin{list}{$\bullet$}
{\setlength{\itemsep}{-0pt}
\setlength{\topsep}{0pt}
\setlength{\labelwidth}{5pt}
\setlength{\leftmargin}{10pt}
\setlength{\parsep}{-0pt}
\setlength{\itemsep}{0pt}
\setlength{\partopsep}{0pt}}}%
{\end{list}}

\makeatletter
\DeclareRobustCommand\onedot{\futurelet\@let@token\@onedot}
\def\@onedot{\ifx\@let@token.\else.\null\fi\xspace}

\def\eg{\emph{e.g}\onedot}

\makeatother

\newcommand{\rpmh}{$\scriptstyle\pm~$}

\definecolor{darkgreen}{RGB}{0,127,0}
\definecolor{darkred}{RGB}{200,0,0}

\begin{document}

\maketitle
\thispagestyle{empty}
\pagestyle{empty}


\input{sections/abstract}
\input{sections/intro}
\input{sections/related}
\input{sections/formulation}
\input{sections/approach}
\input{sections/experiment}
\input{sections/conclusion}

\renewcommand*{\bibfont}{\footnotesize}
\printbibliography 



\end{document}

%% file: sections/abstract.tex
\vspace{-20pt}
\begin{abstract}
We introduce SPOT, an object-centric imitation learning framework. The key idea is to capture each task by an object-centric representation, specifically the SE(3) object pose trajectory relative to the target. This approach decouples embodiment actions from sensory inputs, facilitating learning from various demonstration types, including both action-based and action-less human hand demonstrations, as well as cross-embodiment generalization.  
Additionally, object pose trajectories inherently capture planning constraints from demonstrations without the need for manually-crafted rules. 
To guide the robot in executing the task, the object trajectory is used to condition a diffusion policy. We systematically evaluate our method on simulation and real-world tasks. In real-world evaluation, using only eight demonstrations shot on an iPhone, our approach completed all tasks while fully complying with task constraints. Project page: \textcolor{blue}{https://nvlabs.github.io/object\_centric\_diffusion}


\end{abstract}

%% file: sections/intro.tex
\section{Introduction}

Learning from demonstrations is an effective method for acquiring complex robotic manipulation skills. This approach simplifies learning vision-based manipulation policies by training directly on demonstration data, thus avoiding the need for intricate hand-crafted rules or laborious robot self-exploration in real-world settings. However, developing a structural representation that captures task-relevant information and remains resilient to environmental perturbations continues to be a significant challenge.

Previous work~\cite{reed2022generalist, jang2022bc, ahn2022can}  has utilized raw RGB inputs for end-to-end visuomotor policy training. However, the absence of explicit 3D representation hinders generalization to various viewing angles and out-of-distribution spatial configurations. Instead, RVT~\cite{goyal2023rvt} and PerAct~\cite{shridhar2023perceiver} use the entire 3D scene as input represented by point clouds or 3D volumes. While these methods consider the overall 3D scene structure, they include excessive redundant information,  necessitating extensive demonstration data collection. Additionally, the tight coupling of observation and action pairs complicates deployment across different camera setups, or learning from action-less data like human hand demonstrations.

Object-centric methods address these problems by extracting object-centric information from visual observation, such as object detection~\cite{zhu2023viola}, 6D object poses~\cite{wen2022you}, and 2D or 3D flows~\cite{zhu2024vision,xu2024flow,yuan2024general}, to serve as inputs for downstream policies. By leveraging visual foundation models, these methods exhibit impressive generalization capabilities in new environments and demonstrate strong data efficiency. Nevertheless, most works do not model the complete object pose trajectories or focus only on the last-inch manipulation policy. This oversight necessitates extra human effort to manually create rules for constrained planning, especially in tasks where intermediate actions are crucial for success. For example, in tasks like pouring water or serving a plate, the container must remain upright throughout to prevent spillage. Such constraints are non-trivial to automatically learn with existing object-centric frameworks. 

\begin{figure}[t]
    \centering
    \includegraphics[width=1.0\linewidth]{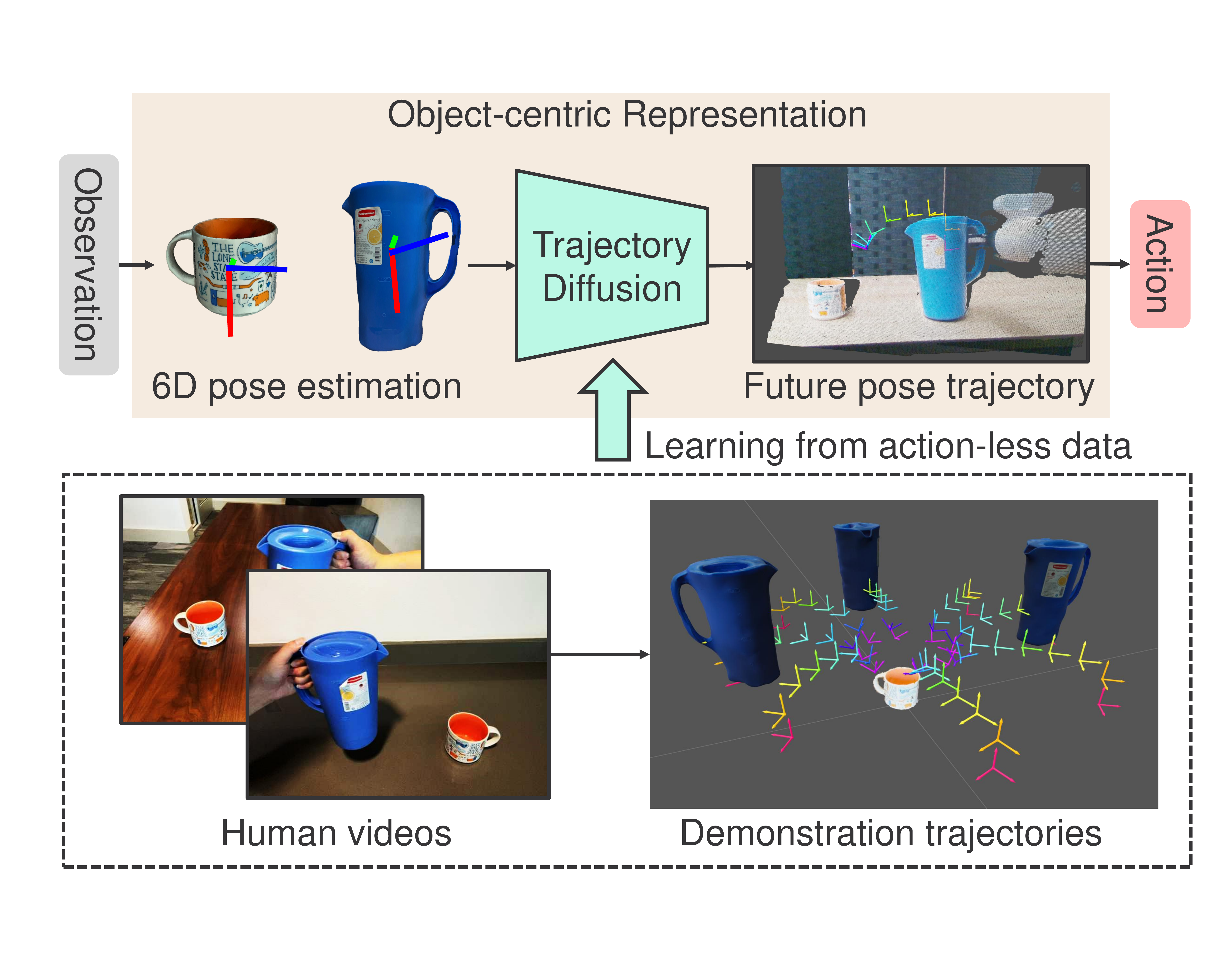}
    \vspace{-35pt}  
    \caption{
        We present SPOT, an imitation learning method that leverages object pose trajectories as an intermediate representation. Given the observation, our framework estimates the object’s pose relative to the target, predicts its future path in SE(3), and derives an action plan accordingly. Our diffusion model is trained on demonstration trajectories extracted from videos without needing action data from the same embodiment.
    }\label{fig: teaser}
    \vspace{-5pt}  
\end{figure}

To address these issues, we introduce SPOT, an object-centric diffusion policy framework that combines diffusion policies and an object-centric representation for manipulation. As shown in Fig.~\ref{fig: teaser}, it uses the SE(3) object-centric trajectory relative to the target as the representation. This approach decouples embodiment action from sensory input, facilitating learning from diverse embodiments and demonstration data, whether action-based or action-less. In particular, in contrast to the standard diffusion policy~\cite{chi2023diffusion,ze20243d} which directly produces robot end-effector actions, our diffusion model learns to synthesize target SE(3) object-centric pose trajectories along the horizon, conditioned on the current observed object pose. The training data is obtained by extracting such object-centric information from demonstration videos. For robot execution, the action is computed from the task space to transport the object following the target synthesized trajectory. By recurrently switching between future trajectory synthesis and trajectory following, we achieve closed-loop control with enhanced robustness under dynamic uncertainty, such as in-hand object motion during grasping or placement. Unlike previous methods that focus solely on the last inch of manipulation, our approach trains on the pose trajectory data over the complete horizon, eliminating the need for manually-crafted global planning rules before the last-inch stage. It also leverages data scaling laws when large amounts of demonstration data are available while maintaining data efficiency in few-shot demonstration scenarios. 

We evaluate the approach on RLBench~\cite{james2020rlbench}, where SPOT outperforms all baselines in high-precision and long-horizon tasks in a challenging single-camera setting. In real-world tasks, with just 8 demonstrations shot on an iPhone, the resulting policies complete all tasks while fully complying with task constraints. We will make all datasets and code available so others can reproduce and expand upon this work. In summary, our framework has the following desired properties:
\begin{myitem}
    \item \textbf{Flexibility on diverse demonstration data.} Our framework can learn from action-based robotic demonstrations and action-less data from human demonstrations. This flexibility enables easier data collection using readily-accessible setups, such as a handheld iPhone, without the need for robotic arms, as our experiments demonstrate.
    \item \textbf{Autonomy on complex task constraints.} Our framework considers the entire sequential object pose trajectory. This approach allows it to automatically learn planning constraints from demonstrations, eliminating the need for manually-crafted rules.
    \item \textbf{Closed-loop feedback.} Our diffusion model offers control over the horizon of action prediction and execution, enabling closed-loop control based on updated observations. This enhances robustness in dynamic situations, like when an object slips during manipulation or placement.
    \item \textbf{Generalization}. Our object-centric representation allows us to ignore task-irrelevant information from the background and thus generalizes robustly to varying lighting conditions, scene setup, and sensory configurations.
    \item \textbf{Language-conditioned multi-task policy}. Our framework inherits the benefits of diffusion models such that it can condition on language or other modalities to deploy on various tasks with a single set of weights, with negligible performance change compared to per-task policy training.
\end{myitem}


  

%% file: sections/related.tex
\section{Related work}



\noindent\textbf{End-to-end Imitation Learning}. Significant advancements have been achieved in end-to-end imitation learning~\cite{ze20243d,chi2023diffusion,shridhar2023perceiver,goyal2024rvt,du2024learning,fu2024mobile}, which involves directly mapping raw sensory observations to action predictions. This field can be broadly divided into two categories: discriminative models and generative models. The discriminative model aims to directly regress the action output given the observations from the collected demonstration data~\cite{shridhar2023perceiver,goyal2024rvt}. Representative work PerAct~\cite{shridhar2023perceiver} encodes the RGBD image of the entire scene as a 3D volume. With additional language conditioning, it achieves promising results with a low number of demonstrations. RVT~\cite{goyal2023rvt,goyal2024rvt} further enhances its efficiency by projecting the 3D scene cloud into orthogonal virtual views. However, the discriminative model often struggles with challenging action prediction-tasks with multimodal distributions, sequential actions, or high precision. Recent works~\cite{chi2023diffusion,ze20243d,ke20243d,yan2024dnact} leverage diffusion models to learn an action distribution and synthesize action conditioned on the raw sensory input. Despite promising results, this work is limited by the need for extensive demonstration data, limited cross-embodiment generalization, and difficulty adapting to new scenarios.

\noindent\textbf{Imitation Learning from Video}.
Recent advancements have shifted from relying on explicit robot action data to learning from visual demonstrations, which can be performed by either humans or robots~\cite{bahl2022human,du2024learning,pari2021surprising,nair2022r3m,sivakumar2022robotic,kurutach2018learning,karnan2022adversarial,baker2022video,ryu2022equivariant,sharma2019third,wen2022you}. Some recent approaches~\cite{vecerik2024robotap,sieb2020graph} extract object state information from video demonstrations and frame the manipulation problem as action replay via low-level visual servoing primitives. Although these methods show promise, reasoning solely on action primitives like picking and placing ignores intermediate actions, preventing applications to tasks where intermediate action needs to satisfy certain constraints. Other methods~\cite{wen2023any, xu2024flow,lin2024flowretrieval} use optical flow or point tracking as intermediate representations from video demonstrations, with additional policy networks to output actions. However, these policies rely on robot action demonstrations tied to the same embodiment, restricting cross-embodiment transfer. Another line of work synthesizes videos of desired task executions and determines actions by learning an inverse-dynamics model~\cite{du2024learning} or extracting target object poses using optical flow~\cite{ko2023learning}. Instead of synthesizing videos, which can be computationally intensive and noisy, we directly generate the object pose trajectory end-to-end.


\noindent\textbf{Object-Centric Manipulation}. A line of research leverages object-centric representations to reason about visual scenes in a modular way to promote policies prioritizing task-relevant factors while reducing the influence of irrelevant visual distractions~\cite{zhu2023learning,wen2022you}. Varying forms of object centric representations have been explored, including object pose~\cite{wen2022you,migimatsu2020object,morgan2021vision,morgan2022complex,vitiello2023one}, object detection/segmentation~\cite{valassakis2022demonstrate,devin2018deep,wang2019deep, zhu2023viola}, point tracking~\cite{xu2024flow,yuan2024general,huang2024rekep,zhu2024vision,zhang2023flowbotpp,seita2023toolflownet,vecerik2024robotap,gao2021kpam,sundaresan2023kite}, neural implicit fields~\cite{simeonov2022neural,urain2023se,weng2023neural,jauhri2023learning}, or object-centric latent features extracted from visual foundation models~\cite{di2024dinobot,di2024keypoint}. Among these efforts, it is common to combine object-centric perception with conventional planning~\cite{morgan2021vision}, or only focus on learning complex policies for last-inch manipulation~\cite{wen2022you,valassakis2022demonstrate,di2024dinobot}. These approaches often depend on manually-programmed planning constraints for tasks where intermediate sequential actions are critical~\cite{gao2021kpam,manuelli2019kpam,huang2024rekep}, making it difficult to scale to multiple tasks with a single policy and set of hyperparameters. Moreover, the typical treatment of a constant rigid transform between gripper and object results in open-loop execution and lacks reactivity to dynamic uncertainties~\cite{sundaresan2023kite,gao2021kpam,manuelli2019kpam}. Our framework inherits the prominent generalization properties of object-centric representations by building upon recent visual foundation models, including object pose estimation and segmentation. We additionally automate task learning for the complete episode and minimize human intervention by reasoning over sequential object pose trajectories. Closed-loop feedback during robot execution further enhances robustness to dynamic uncertainties when dealing with high-precision tasks.


%% file: sections/formulation.tex
\begin{figure*}[t]
    \centering
    \includegraphics[width=1.0\textwidth]{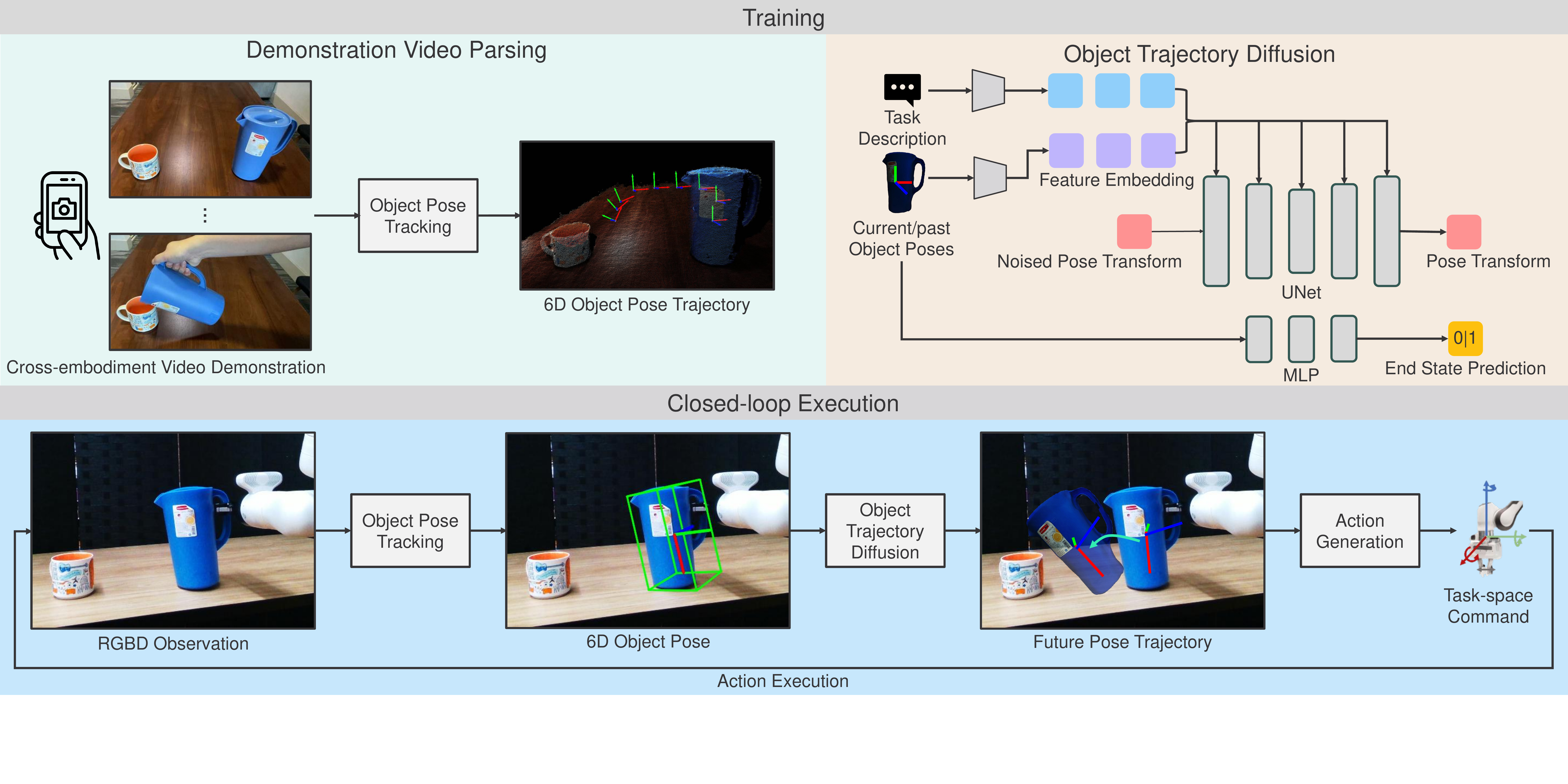}
    \vspace{-40pt}  
    \caption{
        \textbf{Overview.} During training, we extract object pose trajectories from demonstration RGBD videos (\eg., collected with an iPhone), which are independent of the embodiment. Using these extracted trajectories, we train a diffusion model to generate future object trajectories and determine task completion based on current and past poses. During task execution, the task-relevant object is constantly tracked, and its pose is forwarded to the trajectory diffusion model to predict the object's future trajectory in SE(3) that leads to task accomplishment. Finally, we convert the generated trajectories into embodiment-agnostic action plans for closed-loop manipulation.
    }
    \label{fig: framework}
\end{figure*}

\section{Problem Setup}


We aim to learn a policy $\pi$ that maps the observation $O_t$ to the robot’s action $A_{t:t+h}$, where $h$ is the prediction horizon. A handful of demonstrations are provided for learning, which can be performed by either a robot or a human and recorded as RGBD videos. The policy focuses on prehensile manipulation tasks involving rigid objects. The observation $O_t$ represents the robot’s sensory data, including RGBD images. For the action space, we utilize positional control, meaning $A_t \in SE(3)$ is specified by end-effector poses. In the case of a single multi-task policy, task descriptions are provided to specify task-relevant objects and desired goals. 



%% file: sections/approach.tex
\section{Approach}
Fig.~\ref{fig: framework} shows our approach, which uses object pose trajectories as an intermediate representation for manipulation policy learning. We extract object trajectories from demonstration videos that can be performed by either a robot or a human. With the extracted object pose trajectories, we train a diffusion model to predict future object trajectories conditioned on the object pose history. For multi-task settings, the model can condition on a task description embedding. The generated trajectories are then converted into embodiment-agnostic action plans for closed-loop manipulation. Unlike end-to-end methods that tie grasp poses to subsequent motion, our object-centric approach is hardware-independent, allowing seamless integration with any grasp selection technique.

\subsection{Demonstration Video Parsing}
\label{sec: pose_extraction}

We opt for object pose trajectories as the object-centric representation, which depicts the full SE(3) state information of rigid objects throughout the episode. Previous methods that derive SE(3) poses from dense point flow~\cite{vecerik2024robotap} can introduce redundancy and noise; instead, directly reasoning about object poses enhances accuracy, as demonstrated in our experiments. Additionally, the compact nature of object poses improves compatibility with diffusion model training and boosts learning efficiency. To learn a policy from the demonstration video, the key step is to extract task-relevant objects and their pose trajectories.

Without loss of generality, we consider a single-arm manipulation scenario where only one object is manipulated at any given time. As a result, the task involves a set of objects $V = \{v_a, v_b\}$, consisting of a graspable source object $v_a$ and a target object $v_b$. For example, in a water-pouring task, the graspable object might be a kettle, and the target object might be a mug. Given the demonstration video with length $l$, the goal is to extract pose trajectory $\{\hat{T}_a\}_0^l$ and $\{\hat{T}_b\}_0^l$ corresponding to object $v_a$ and $v_b$, respectively. These trajectories describe the task-specific object-centric motions in 3D and thus implicitly encode the planning constraints throughout the task, including intermediates constraints such as keeping the kettle upright to prevent spilling until it approaches the mug. In practice, we apply a zero-shot 6D object pose estimation method~\cite{wen2024foundationpose} to extract the pose trajectory from the video (Sec.~\ref{sec: pose_tracking}) for both the source and target objects. 
We then treat the source and target objects as two nodes in a scene graph and convert all source object poses into the target object’s frame $\hat{\tau} = \{\hat{T}^a_b\}_0^l$. This transforms multiple demonstration trajectories into a canonical space regardless of their absolute configurations or backgrounds, allowing demonstration data to be collected in various environments.

Training directly on the dense, continuous pose trajectory can be inefficient and unstable due to noisy, shaky motions from human hands in the video. To address this, we adopt the keyframe selection approach described in PerAct~\cite{shridhar2023perceiver}, which is summarized as follows: a frame is added if the relative velocity is zero (indicating a change in direction) or if it exceeds a certain distance threshold from the previous keyframe, measured in both translation and rotation.

\subsection{Object Trajectory Diffusion}
\label{sec: traj_modeling}


Our method decouples observation-action pairs in visuomotor policy learning by employing object pose trajectory as an intermediate representation. We model $p(T_{t:t+h} | T_{t})$ to infer a future object pose trajectory  $T_{t:t+h} \in SE(3)$ that accomplishes the task, from the current object pose $T_t \in SE(3)$. 

Specifically, we use the extracted object trajectories to train a task-dependent model that can synthesize the object pose trajectory.
Our framework is built upon diffusion policy~\cite{chi2023diffusion}, which leverages a denoising diffusion probabilistic model (DDPM) to capture the conditional action distribution $p(A_t|O_t)$:
\begin{align}
    A_t^{k-1} = \alpha\left(A_t^k - \gamma \epsilon_\theta(O_t, A_t^k, k) + \mathcal{N}(0, \sigma^2 I)\right)
    \label{eq: ddpm}
\end{align}
The process starts with ${A}_t^K$ sampled from Gaussian noise and is iterated $K$ times. Here, $(\alpha, \gamma, \sigma)$ are the parameters of the denoising scheduler. This process is also called the reverse process and is detailed in previous work~\cite{ho2020denoising}. The score function $\epsilon_\theta$ is trained by the loss function
\begin{align}
\mathcal{L} = \text{MSE} \left( \epsilon_k, \epsilon_\theta\left(O_t, A_t^0 + \epsilon_k, k\right)\right)
    \label{eq: dp_loss}
\end{align}

In our framework, we model $p(T_{t+1} \mid T_t)$ as the primary component for policy learning. Rather than predicting future actions, we synthesize the object pose trajectory for task progression. Additionally, our diffusion process is conditioned on the historical pose until now rather than raw sensory input. To adapt this method to our framework, we replace the observation input $O_t$ with the current object pose $T_t$. Furthermore, to obtain the object pose trajectory, $A_t$ is replaced by the transformation $T_{t+1}$. Accordingly, we modified Eq.~\eqref{eq: ddpm} to
\begin{align}
    T_t^{k-1} = \alpha\left(T_t^k - \gamma \epsilon_\theta(T_{t}, T_{t+1}^k, k) + \mathcal{N}(0, \sigma^2 I)\right)
    \label{eq: ddpm2}
\end{align}

The training loss introduced from Eq.~\eqref{eq: dp_loss} now becomes:
\begin{align}
\mathcal{L} = \text{MSE} \left( \epsilon_k, \epsilon_\theta \left( T_{t}, T_{t+1}^0 + \epsilon_k, k \right) \right)
    \label{eq: dp_loss2}
\end{align}

\noindent  \textbf{Pose Feature Encoder.}
Inspired by previous work~\cite{chi2023diffusion}, we employ a lightweight MLP network to encode object poses into compact representations. This network features a three-layer MLP, LayerNorm~\cite{ba2016layer} layers, and a projection head, which projects the final features into a 64-dimensional vector.

\noindent  \textbf{End State Prediction.} To control the gripper state upon task completion, we employ a separate MLP network as a binary classifier to indicate task completeness. The classifier takes the current object pose as input and determines whether the rollout has reached its end state. 
We assign 1 to the end state and 0 to all other states, supervised by binary cross-entropy loss. During deployment, the gripper opens when the prediction is beyond a threshold of $0.95$.

\noindent  \textbf{Multi-task Language Conditioning.} Thanks to the flexibility of diffusion models, we can effortlessly adapt to single-model, multi-task settings by conditioning on other multimodal data, such as language. Specifically, task descriptions are fed into a pretrained CLIP~\cite{radford2021learning} model to obtain sentence embeddings, following previous work~\cite{shridhar2023perceiver, goyal2023rvt, goyal2024rvt}. These feature embeddings are then concatenated with each object pose as input to the model.

We use DDIM~\cite{song2020denoising} as the noise scheduler. The scheduling is set to 100 timesteps during training and 10 timesteps during inference. The number of training epochs is 3000, and the batch size is 128 for all the baselines.



\subsection{Closed-loop Policy Execution}
\label{sec: close_loop_execution}

To derive the closed-loop manipulation policy $\pi$ from our object-centric representation, we need two key connections: one from observation to object pose, $O_t \mapsto T^{obj}_{cam}$, and another from pose trajectory to action plan, $T_{t:t+h} \mapsto A_{t:t+h}$. For the first mapping, we utilize off-the-shelf object pose tracking. For the second mapping, the action plan is derived from the synthesized object trajectory. The complete framework consists of three main components: \textit{Object Pose Tracking}, which provides real-time object poses at each step; \textit{Trajectory synthesis}, which takes these poses as input and outputs the future object trajectory for task progression; and finally, \textit{Action plan generation}, which derives the action plan from the synthesized object trajectory.

\noindent  \textbf{Object Pose Tracking.}
\label{sec: pose_tracking} 
We utilize FoundationPose~\cite{wen2024foundationpose}, a foundation model for 6D object pose estimation and tracking, which can be applied to novel objects in a zero-shot setting. It tracks all the task-relevant SE(3) object poses throughout the policy rollout to provide the latest feedback to the controller.

In addition to RGBD observations, the tracker requires a 3D object model and 2D detection for the first frame. We obtain object scans using an iPhone equipped with a camera and LiDAR, and object bounding boxes by YOLO-World~\cite{cheng2024yolo}. Similarly to demonstration data processing,  we convert the estimated source object pose into the target object’s coordinate frame before feeding it into the trajectory diffusion model.


\noindent  \textbf{Trajectory synthesizing.}
At test time, we employ the trained diffusion model (Sec.~\ref{sec: traj_modeling}) to generate future object trajectories that can accomplish the task. At each step, the model uses the previous (if the observation horizon is greater than 1) and current object poses as inputs to predict future object pose trajectories. We aim to achieve long-horizon planning while remaining reactive to unexpected dynamic uncertainties, such as in-hand object slipping or moving target objects. To this end, we integrate the object pose trajectories synthesized by the diffusion model with receding horizon control~\cite{mayne1988receding} to ensure robust action execution. Specifically, the model forecasts future object pose trajectories for $N$ steps, from which we select $K$ steps to execute actions. This design enables our approach to achieve closed-loop control by incorporating updated observations from the object pose tracker.

\noindent  \textbf{Action plan generation.} The synthesized trajectory indicates the desired object poses to complete the task. To convert the pose trajectory into an executable action plan, we transform the future object poses into end effector subgoals using the transformation $T^{obj}_{EE} \in SE(3)$ from the object pose to the end effector pose. Given the extrinsic camera calibration, we can derive the transformation $T^{obj}_{EE}=T^{EE}_{cam}\cdot T^{cam}_{obj} \in SE(3)$ from the object pose to the end effector pose. Consequently, the action can be derived from the predicted object pose by $A_{t:t+h} = T^{obj}_{EE} \cdot T_{t:t+h}$. The action plan is constructed by following these end-effector subgoals. The policy continues to run until the episode ends and changes the gripper state, as signaled by the end-state prediction.

%% file: sections/experiment.tex
\section{Experiments}
Our experiments aim to address the following research questions: 1) Is our pose trajectory representation effective for capturing task-related 3D information? 2) Can our object-centric approach learn from action-less data and bridge the embodiment and environmental gap between training and testing? 3) Can our object-centric approach generate trajectories that comply with task-specific motion constraints? 4) Can we democratize demonstration data collection with a minimal hardware setup?

\subsection{Experimental Setup}

\noindent  \textbf{Simulation setup.}
We evaluate baselines on RLBench~\cite{james2020rlbench}, a standard multi-task manipulation benchmark adopted by previous studies~\cite{shridhar2023perceiver, goyal2023rvt, goyal2024rvt, ke20243d}. The task and the robot are simulated using CoppeliaSim~\cite{rohmer2013v}. A Franka Panda robot equipped with a parallel jaw gripper is utilized for all tasks. Each task is accompanied by a language description and includes 2 to 60 variations, where the task-relevant objects are in different initial/goal configurations or appearances. We chose 13 manipulation tasks that excluded non-prehensile tasks and tasks with articulated objects, as they are beyond the scope of this work. Following previous work~\cite{shridhar2023perceiver, goyal2023rvt}, all methods are trained on the same set of 100 demonstrations and tested on 25 unseen task configurations.

\noindent  \textbf{Real-world setup.}
To validate learning from videos in unstructured settings without specific environmental or hardware requirements, we recorded demonstration videos in everyday environments using consumer-grade devices. Each task demonstration was performed by a human without robotics hardware, and filmed in various locations such as the kitchen, living room, or office, ensuring no overlap with the robot’s testing environment. This results in diverse visual backgrounds, lighting conditions, and camera setups that are different from those used during robot testing. The videos were captured with an iPhone equipped with RGB and LiDAR sensors, and all manipulations were performed by human hands. For each task, 8 demonstrations were collected for training, and 10 trials were conducted for each method during evaluation. All real-world experiments were conducted using the same hardware setup: a robot equipped with a Kinova Gen3 7-DoF arm and a forward-facing Intel RealSense D415 RGBD camera mounted at the left shoulder of the robot, which was extrinsically calibrated.

\subsection{Simulation Results}
The evaluations are displayed in Table~\ref{tab: rlbench}. This comparison addresses question (1) by demonstrating the effectiveness of our method across a diverse set of tasks, particularly in the challenging single-view setting.

\noindent  \textbf{Implementation details.}
To collect demonstration trajectories, we use the opening or closing of the gripper as signals to start or terminate a trajectory. Each trajectory corresponds to an episode for single-stage tasks or a stage for multi-stage tasks. For multi-stage tasks, in addition to language embedding, the object trajectory diffusion model is additionally conditioned on a one-hot encoding of the task stage. The next task stage is automatically triggered given the predicted end state (Sec.~\ref{sec: traj_modeling}).
Our approach focuses on generating post-grasp trajectories, relying on pre-generated grasp poses (derived from the demonstration) to grasp the object. Subsequently, actions are executed based on a synthesized pose trajectory. 
For object pose tracking, we reconstruct the object mesh from the training demonstration sequence using BundleSDF~\cite{wen2023bundlesdf} and obtain the first frame object mask with SAM6D~\cite{lin2024sam}.


\noindent  \textbf{Baselines.}
We compare with two state-of-the-art 3D-based imitation learning methods: RVT2~\cite{goyal2024rvt} and 3D Diffuser Actor~\cite{ke20243d}. Both train language-conditioned, multi-task policies. We use their officially-released implementations for retraining on the set of considered tasks.
Notably, RVT2 requires a multi-camera setup (4 in this case), whereas 3D Diffuser Actor supports single-view, the same as our setup. 

\input{tables/rlbench_icra}

\noindent  \textbf{Analysis.}
Our method achieves performance comparable to the multi-view approach RVT2, even in a single-camera setting where occlusion often occurs. Additionally, our method excels in high-precision tasks such as {\fontfamily{qcr}\selectfont {Insert Peg}} and {\fontfamily{qcr}\selectfont {Place Cups}}, outperforming other baselines. By leveraging a pose trajectory representation, our method captures the essential spatial relationships between the task-relevant objects, as well as the intermediate transformations during the process. Furthermore, our method surpasses other baselines in long-horizon tasks like {\fontfamily{qcr}\selectfont {Stack Blocks}} and {\fontfamily{qcr}\selectfont {Stack Cups}}, as the object-centric representation naturally decomposes long-horizon, multi-object task learning into a series of sub-policies. However, our method struggles with thin objects (the stick in {\fontfamily{qcr}\selectfont {Drag Stick}}), small objects occluded by the gripper (the bulbs in {\fontfamily{qcr}\selectfont {Screw Bulb}}),  or a mix of objects with different symmetries (the shape toys in {\fontfamily{qcr}\selectfont {Sort Shape}}).

\subsection{Real-world Results}
The evaluations for the four tasks are shown in Figure~\ref{fig: real}. This comparison addresses questions (2)-(4) by determining whether our method can transfer 3D trajectories captured in cross-embodiment videos (human demonstrations) and successfully meet the tasks' motion constraints through automatic learning under minimal data collection efforts.

\noindent  \textbf{Tasks and evaluation protocol.}
We designed four tasks to evaluate policy performance: 1) {\fontfamily{qcr}\selectfont {mug-on-coaster}}: placing a mug on a coaster; 2) {\fontfamily{qcr}\selectfont {plant-in-vase}}: inserting a cactus plant into a vase; 3) {\fontfamily{qcr}\selectfont {pour-water}}: using a kettle to pour water into a mug; 4) {\fontfamily{qcr}\selectfont {put-plate-into-oven}}: placing a plate with food into an oven. 
These four tasks involve various challenges commonly considered in robot manipulation. In particular, Task 1 represents the commonly considered pick-and-place task while requiring a reasonable level of precision due to the small size of the coaster. Task 2 requires even higher precision due to the low tolerance between the plant base and the vase. The last two tasks additionally require satisfying the object’s intermediate motion constraints throughout the episode: the kettle and plate must stay upright, and the content should stay inside/on throughout the entire episode. Regarding the task outcome, other than task success, we use \textit{Placing failure} to indicate that the robot fails to deliver the object to the goal position. \textit{Task constraint failure} describes a scenario where the robot successfully delivers the object but fails to meet the task constraints. \textit{Tracking failure} occurs when the object tracker loses track of the object, forcing the rollout to terminate early, as we rely on object tracking to infer the action plan.

\begin{figure}[t]
    \centering
    \includegraphics[width=0.95\linewidth]{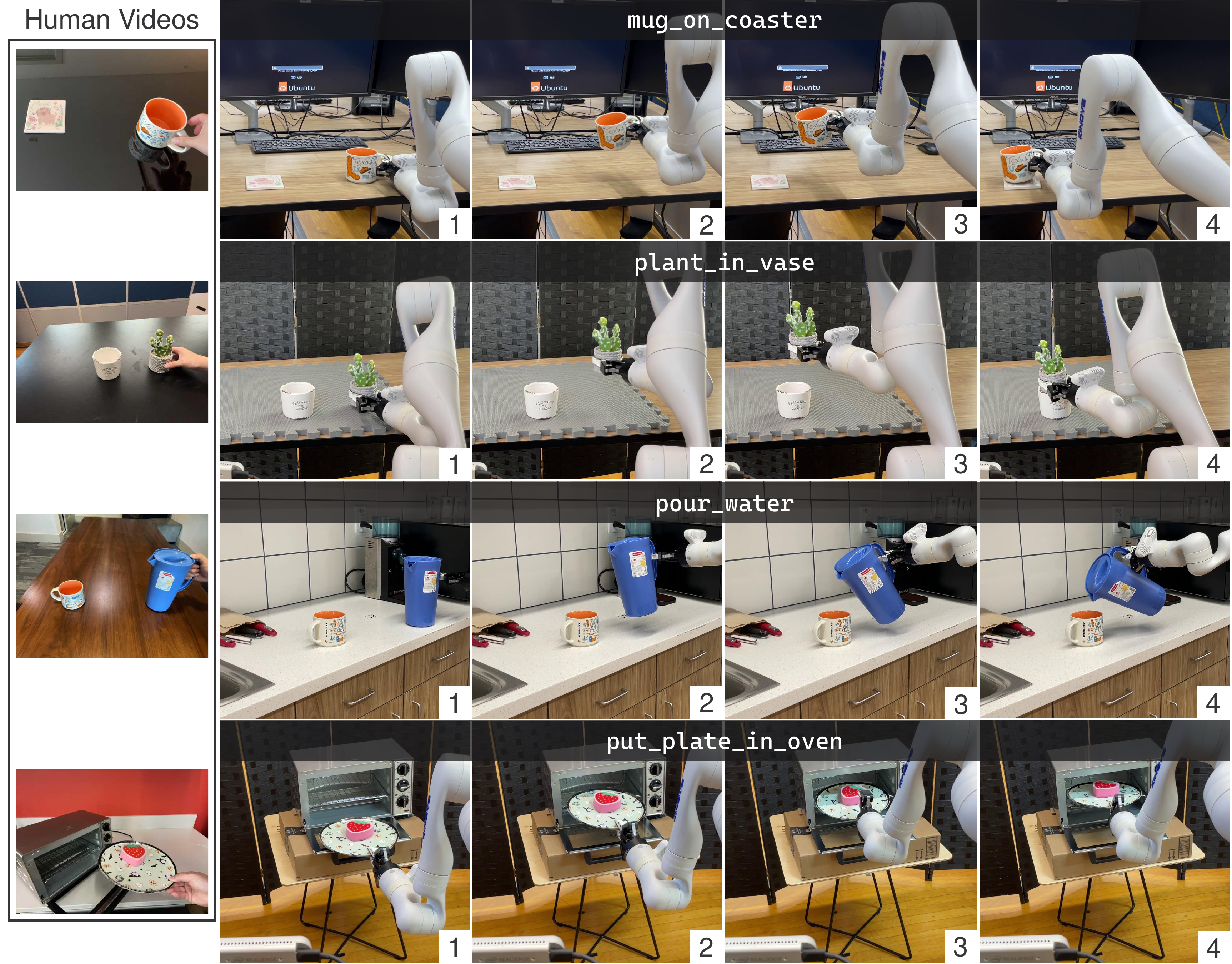}
    \caption{
        \textbf{Real-world Tasks and Qualitative Results.} Demonstration data was collected using an iPhone to record the RGBD video of a human performing the tasks (\textbf{Left}). The robot deploys the trained policy in drastically different environments, lighting conditions, camera perspectives, and object configurations from demonstration time (\textbf{Right}).
    }\label{fig: real}
    \vspace{10pt}  
\end{figure}

\begin{figure}[t]
    \centering
    \includegraphics[width=0.95\linewidth]{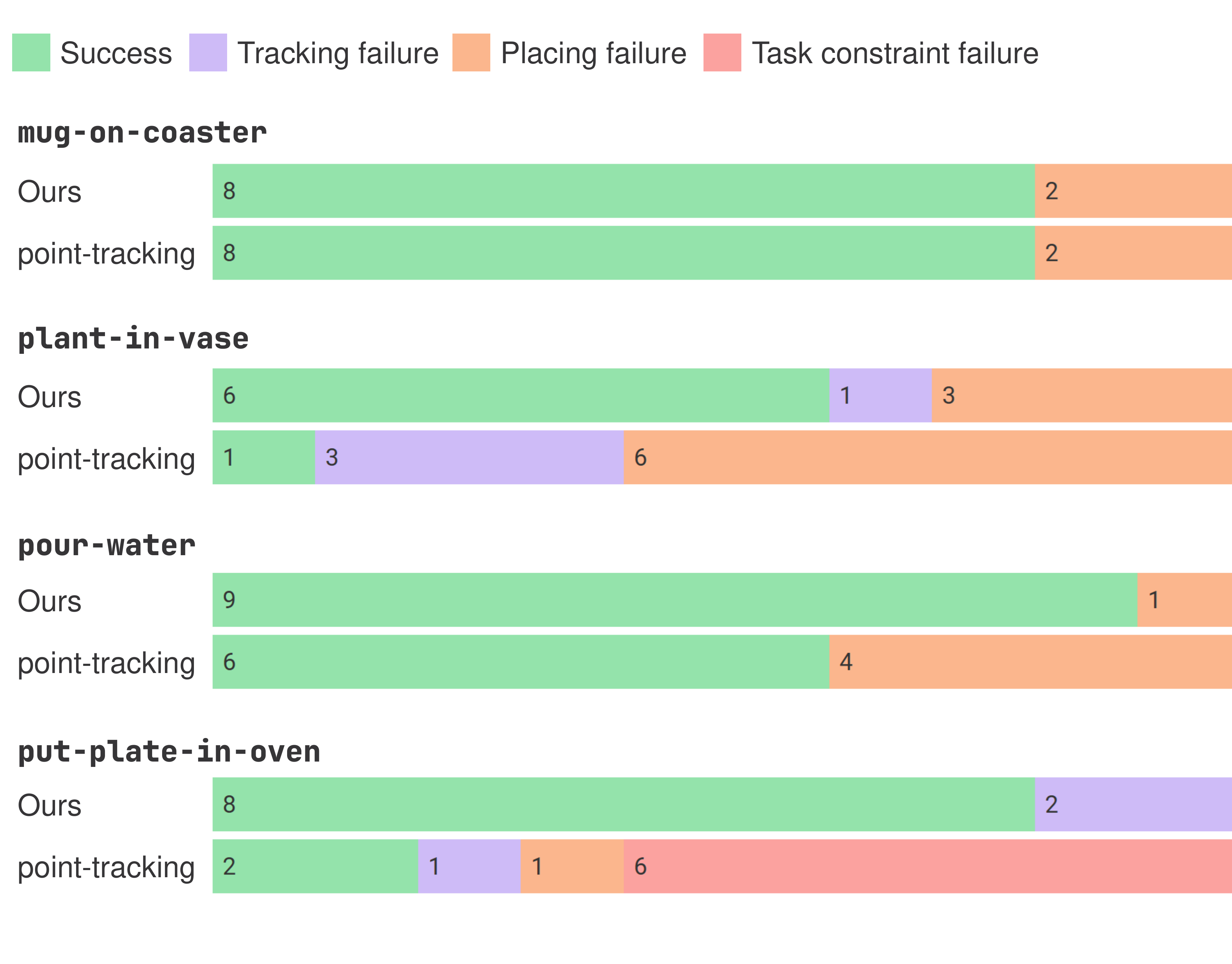}
    \vspace{-11pt}
    \caption{
        \textbf{Real-world Quantitative Results. } Our method outperforms the point-tracking baseline.  We categorize failure modes into (i) tracking failure, (ii) placing failure, and (iii) task constraint failure.
    }
    \label{fig: real}
\end{figure}

\noindent  \textbf{Baselines.}
Learning policies from action-less videos while adhering to task-specific constraints is challenging. Previous works~\cite{wen2023any, vecerik2024robotap, bharadhwaj2024track2act, yuan2024general} exploit point tracking for extracting task-relevant actions from videos. Accordingly, we implement a \textsc{Point-Tracking} baseline, utilizing a heuristic point-tracking method to derive object trajectories from the video. Specifically, we uniformly sample key points from the first frame, track these points using CoTracker~\cite{karaev2023cotracker}, apply RANSAC to extract object transformations throughout the sequence, and use these transformations for learning trajectory synthesis. This baseline allows us to compare our approach with existing methods in modeling intermediate states and implicit task constraints.

\noindent  \textbf{Analysis.}
Without specific task constraints, all evaluation methods perform well in the general pick-and-place task of  {\fontfamily{qcr}\selectfont {mug-on-coaster}}. However, in the {\fontfamily{qcr}\selectfont {plant-in-vase}} and {\fontfamily{qcr}\selectfont {pour-water}} tasks, \textsc{Point-Tracking} struggles to achieve sufficient accuracy for the plant to fit into the vase due to the noisy extracted object trajectory. Additionally, the object pose tracker sometimes fails to track the cactus plant in {\fontfamily{qcr}\selectfont {plant-in-vase}} due to the camera setup and occlusion caused by the robot arm, highlighting a shared limitation of both methods. In the {\fontfamily{qcr}\selectfont {put-plate-into-oven}} task, the robot must keep the plate upright while inserting it into the narrow space inside the oven. Our method outperforms the baseline by producing smooth trajectories while adhering to constraints. In contrast, \textsc{Point-Tracking} often drastically changes orientation, causing content to fall off the plate. 


%% file: tables/rlbench_icra.tex
\begin{table}[t]
    \Huge
    \centering
    \vspace{7mm}
    \caption{\textbf{Simulation Results on RLBench.} All methods use a single set of weights to perform multiple tasks, conditioned on language description. Notably, RVT2 uses a four-camera setup, while 3D-DA and Ours use a single-view setup.
    }     \label{tab: rlbench}
    \resizebox{\columnwidth}{!}{%
   \begin{tabular}{l|ccccccc}
\hline
\rowcolor[rgb]{ .706,  .776,  .906}                & \textbf{Avg}                   & \textbf{Close}                 & \textbf{Drag}                  & \textbf{Insert}                & \textbf{Meat off}              & \textbf{Place}                 & \textbf{Place} \bigstrut[t]\\
\rowcolor[rgb]{ .706,  .776,  .906} \multicolumn{1}{c}{\textbf{Method}} & \textbf{Succ.}                 & \textbf{Jar}                  & \textbf{Stick}                & \textbf{ Peg}                  & \textbf{Grill}                 & \textbf{ Cups}                 & \textbf{ Wine} \bigstrut[b]\\
\hline
RVT2~\cite{goyal2024rvt}       & 76.4                           & \textbf{98.7 \rpmh 2.3}                 & \textbf{98.7 \rpmh 2.3}                 & 44.0 \rpmh 6.9                 & 96.0 \rpmh 0.0                 & 33.3 \rpmh 2.3                 & 92.0 \rpmh 4.0 \bigstrut[t]\\
3D-DA~\cite{ke20243d}          & 54.5                           & 77.3 \rpmh 0.0                 & 92.0 \rpmh 8.0                 & 9.3 \rpmh 2.3                  & 84.0 \rpmh 0.0                 & 4.0 \rpmh 0.0                  & 78.7 \rpmh 4.6 \\
Ours                           & \textbf{79.4}                           & \textbf{98.7 \rpmh 2.3}                 & 80.0 \rpmh 0.0                 & \textbf{78.7 \rpmh 2.3}                 & \textbf{100.0 \rpmh 0.0}                 & \textbf{62.7 \rpmh 6.1}                 & \textbf{100.0 \rpmh 0.0} \bigstrut[b]\\
\hline
\hline
\rowcolor[rgb]{ .706,  .776,  .906}                & \textbf{Put in}                & \textbf{Put in}                & \textbf{Screw}                 & \textbf{Sort}                  & \textbf{Stack }                & \textbf{Stack}                 & \textbf{Turn} \bigstrut[t]\\
\rowcolor[rgb]{ .706,  .776,  .906} \multicolumn{1}{c}{\textbf{Method}} & \textbf{Cupboard}              & \textbf{ Safe}                 & \textbf{ Bulb}                 & \textbf{ Shape}                & \textbf{Blocks}                & \textbf{ Cups}                 & \textbf{Tap} \bigstrut[b]\\
\hline
RVT2~\cite{goyal2024rvt}       & \textbf{69.3 \rpmh 2.3}                 & 89.3 \rpmh 8.3                 & \textbf{86.7 \rpmh 4.6}                 & \textbf{49.3 \rpmh 6.1}                 & 81.3 \rpmh 6.1                 & 60.0 \rpmh 0.0                 & 94.7 \rpmh 4.6 \bigstrut[t]\\
3D-DA~\cite{ke20243d}          & 38.7 \rpmh 2.3                 & 86.7 \rpmh 2.3                 & 38.7 \rpmh 4.6                 & 30.7 \rpmh 2.3                 & 41.7 \rpmh 7.5                 & 30.7 \rpmh 8.3                 & 96.0 \rpmh 0.0 \\
Ours                           & 42.7 \rpmh 4.6                 & \textbf{100.0 \rpmh 0.0}                & 48.0 \rpmh 8.0                 & 32.0 \rpmh 4.0                 & \textbf{94.0 \rpmh 3.4}                 & \textbf{96.0 \rpmh 0.0}                 & \textbf{100.0 \rpmh 0.0} \bigstrut[b]\\
\hline
\end{tabular}%

    }
    \vspace{-20pt}  
\end{table}

%% file: sections/conclusion.tex
\section{Conclusion}
We present SPOT, an object-centric imitation learning method that allows robots to learn from cross-embodiment demonstrations and adhere to task constraints using an intermediate representation of object pose trajectories. Our approach demonstrates robust manipulation capabilities in both simulation and real-world environments, outperforming all baseline methods across a variety of manipulation tasks.

\noindent  \textbf{Limitations.}
SPOT only considers the SE(3) poses of grasped objects, limiting its ability to handle non-prehensile tasks or dynamic tasks that require regulating the robot's speed or force. SPOT depends on 6D pose tracking, which assumes the object is rigid. This method may struggle with small, thin, non-rigid, or deformable objects and could be impacted by significant occlusion. Object tracking requires a reconstructed object mesh, though recent advancements have made this more accessible~\cite{wen2023bundlesdf,wang2024dust3r}.